# Optimizing Train Driving to Minimize the Electricity Cost of an Entire Railway Traffic Mesh using Evolutionary Algorithms

Eduardo Pilo de la Fuente, *Senior Member, IEEE*, Sudip K. Mazumder, *Fellow Member, IEEE,* María Antonia Simón

***Abstract*—This paper presents a procedure to optimize the way trains are driven, which pursues, in addition to fulfilling operational constraints such as admissible speeds or journey durations, the minimization of the cost related to supplying electrical energy to the trains, including the cost of the energy consumption and the cost of the power capacity utilization. This procedure combines: (i) a traffic model that merges the energy and power footprint of each rail service part of the traffic mesh, and (ii) an evolutionary computation framework that enables searching for the optimal way to drive the trains to achieve an optimal traffic mesh. This procedure is applied to a 450 km long section of the high-speed line from Madrid to Barcelona (Spain).**

## I. Introduction

Electric trains are expected to play an essential role in the modernization of transportation systems worldwide, especially in mass transportation systems with journey distances up to 1500 km due to their high energy efficiency and low emissions. A key factor in achieving this efficiency is the electrical interconnection among trains, which, together with regenerative braking (i.e., converting the kinetic energy lost during braking into electrical power, which is supplied to other trains or the grid), enables a richer range of interactions among trains and infrastructure power flows.

However, the best efficiency can only be achieved if the trains' operation (i.e., how all of them are driven and, therefore, when and where power flows) is designed so that the energy and power exchanges improve the performance of the power supply system.

Over the last two decades, system operation optimization has mainly focused on minimizing each train service's energy consumption. This approach has enabled substantial energy savings, resulting in reduced costs and greater efficiency. Nevertheless, one of its main limitations is that it disregards the costs associated with power capacity utilization, as well as any technical limits that depend on it, such as the maximum power a substation can handle. By nature, these aspects do not rely solely on the behavior of an individual train, but rather on the entire traffic mesh, which aggregates all trains using the same infrastructure.

The progressive adoption of Smart Grid technologies in railway power systems will enable more precise control of power flows and, consequently, allow the implementation of new optimization strategies. This paper aims to provide new insights into how to design the operation of electric trains to optimize both energy and power exchanges, thereby improving the overall performance of the power supply system.

In addition to this introduction, this paper includes four sections. Section II describes the state of the art in the optimization of railway operations. Section III describes the optimization procedure proposed in this paper. Section IV discusses the results from a case study of a 450 km-long high-speed line. Finally, Section V summarizes the paper's conclusions.

## II. Optimization of Railway Operation

The optimization of train operation has been extensively addressed in the technical literature, typically focusing on minimizing train energy consumption, evolving from simple mechanical energy minimization for a single train to more complex, systematic approaches that integrate multiple trains, power supply characteristics, and economic factors.

### *A. Mass Rail Transit (MRT) and Subway systems*

MRT and Subway systems, typically powered by DC networks that typically use 750V, 1500V, or 3000V, are characterized by frequent acceleration and braking cycles, making them ideal candidates for regenerative braking energy (RBE) utilization. Thus, various optimization strategies have been proposed to reduce energy consumption and to improve RBE utilization in this type of system.

As described in the review presented in [1], two common approaches for this are: (i) optimizing the way trains are driven (i.e., the sequence of traction forces applied during the journey)

Eduardo Pilo de la Fuente is with Higher Polytechnic School, Universidad Francisco de Vitoria, Ctra. Pozuelo-Majadahonda Km 1,800, 28223, Pozuelo de Alarcón, Madrid., Spain (e-mail: eduardo.pilo@ufv.es).

Sudip K. Mazumder is with the Electrical and Computer Engineering Department, University of Illinois at Chicago, Chicago, IL, 60607, USA (e-mail: mazumder@uic.edu).

Maria Antonia Simón is with Higher Polytechnic School, Universidad Francisco de Vitoria, Ctra. Pozuelo-Majadahonda Km 1,800, 28223, Pozuelo de Alarcón, Madrid., Spain (e-mail: mariaantonia.simon@ufv.es).

Color versions of one or more of the figures in this article are available online at http://ieeexplore.ieee.org.

and (ii) synchronizing the acceleration and braking of several trains. Please note that while both approaches reduce energy consumption, the latter can also reduce the railway system's power peaks by compensating for them through regeneration.

[2] employs Genetic Algorithms (GA) to manage dwell times, aiming to avoid the simultaneous acceleration of too many trains. This results in a smoother traction power load curve and a significant reduction in the peak traction power.

[3] addresses the design of underground rail timetables (Madrid metro Line 3) to synchronize train movements. The model specifically aims to synchronize the braking of trains arriving at a station with the acceleration of trains exiting stations connected to the same electrical section to reduce substation demand.

In [4], ATO speed profiles are designed to minimize net energy at substations for the Madrid underground. It considers train synchronization through a network model that evaluates the recovery coefficient, which characterizes the proportion of RBE used by other motoring trains within the DC network.

[5] proposes a cooperative train control model where an accelerating train reuses the regenerative energy from a braking train on the opposite track. The optimization variables include the departure time of the accelerating train to maximize temporal and spatial overlap with the braking train, thereby reducing the substation's "practical energy consumption".

[6] explores cooperative control in electric subways by taking two trains within the same section as an example. It concludes that energy consumption can be reduced by up to 19.2% by adding a specific traction process to the tracking train, precisely synchronized with the braking phase of the front train, resulting in a five-mode driving strategy.

Focusing on metro systems under DC traction networks, Chen et al. (2021) presents a method for cooperative eco-driving. It uses a multi-phase dynamic programming algorithm to obtain optimal speed profiles, in which the tracking train is optimized to absorb RBE, and the preceding train is optimized to actively generate RBE during the follower's acceleration phase, thereby smoothing power spikes at the substation level.

[8] investigates urban rail transit and introduces the overlap current method. The goal is to maximize current compensation at the substation level by adjusting dwell times to synchronize the acceleration of one train with the braking of another, significantly increasing RBE utilization compared to traditional time-overlap methods and, indirectly, shaving power peaks by ensuring current demand is balanced by current regeneration.

### *B. High-Speed Railways*

High-speed trains operate over longer distances where aerodynamic resistance and checkpoint constraints become critical. From an electrical point of view, AC substations are normally reversible -no rectification is required in them-, which makes train synchronization less impactful in terms of energy saving.

[9] focuses on optimizing manual driving strategies for high-speed trains to minimize traction energy consumption. By means of a Genetic Algorithm (GA) combined with a high-fidelity simulator, a set of guiding commands for drivers (e.g., holding speeds and coasting points) is obtained. The method was tested on the Spanish high-speed line Madrid–Barcelona, achieving measured average energy savings of 18% to 23%.

[10] develops an energy-efficient driving strategy specifically for high-speed trains, based on data from the Beijing-Shanghai high-speed line. The model incorporates discrete control gears and checkpoint constraints (i.e, intermediate points where the train must arrive at a prescribed time to ensure punctuality over long distances). The optimization focuses on reducing traction energy consumption using Karush-Kuhn-Tucker (KKT) conditions to determine optimal regime switching.

[11] proposes an energy-saving driving strategy for high-speed trains using the Proximal Policy Optimization (PPO) deep reinforcement learning algorithm. It uses parameters from the Wuhan-Guangzhou high-speed line. The optimization objective is defined by a reward function that balances punctuality and energy consumption, aiming to minimize the total traction work.

[11] introduces an advanced cruising-coasting strategy tested on both metro and high-speed railway lines. The high-speed case study utilizes real-world data from a Spanish high-speed line (Calatayud to Zaragoza). The objective is to minimize total energy consumption and journey time by identifying optimal trajectories through NSGA-II.

While focused on emergency self-rescue, [12] optimizes the trajectories of Electric Multiple Units (EMUs) trains, the standard rolling stock for high-speed rail. The model optimizes speed curves to minimize net energy consumption (including traction and auxiliary systems) subject to limited on-board battery capacity during power outages on the catenary.

In high-speed railways, operational optimization focuses almost exclusively on energy efficiency and punctuality as the primary objectives, rather than on reducing power peaks at substations.

### *C. Heavy Freight Rail Systems*

Compared to urban rail (DC) or high-speed rail (AC), heavy freight rail studies (typically AC) are notable for their extreme power demand, which leads to optimization strategies that differ slightly, in which voltage stability may be a concern.

[13] establishes a "train-track-power" (TTP) grid model for freight multitrain operation based on the Shuohuang Railway in China. It explicitly focuses on the scenario in which the traction action of one train and the braking action of another overlap, allowing the transfer of RBE. The optimization coordinates the running trajectories of multiple trains to transform "random loads into planned loads," ensuring that accelerating trains absorb the energy generated by braking trains within the same power supply section.

[14] presents dynamic modeling for multi-train freight systems. Like the previous study, it optimizes the allocation and utilization of RBE by synchronizing the speed profiles of multiple trains. It uses a large-scale adaptive algorithm (LA-MOCSO) to solve the high-dimensional problem of coordinating multiple trains over long-distance freight routes.

### *D. Billing Optimization*

[15] analyzes commuter rail (Malaga) on DC networks (3000V) and optimizes the contracted power capacity. Constraints involve Spanish electricity tariffs, failure rates, and traffic density. The method uses pseudo-static simulations to determine the optimal capacity to reduce the total electricity bill, specifically addressing power peaks by accounting for economic penalties for exceeding contracted limits.

## III. Proposed Optimization Procedure

### *A. Overview*

This paper presents an optimization model that combines the following elements (see Fig. 1): (i) a single train simulation module that calculates the movement and the power consumption of each train, (ii) the traffic mesh generator, which determines all the trains that circulating at each simulation step, their position and power consumption, (iii) the objective function calculation, that determines the total cost of the electricity used by all the trains considered in the traffic mesh, and (iv) an evolutionary algorithm that is able to explore different ways of movement of the trains so that the cheapest ones can be discovered.

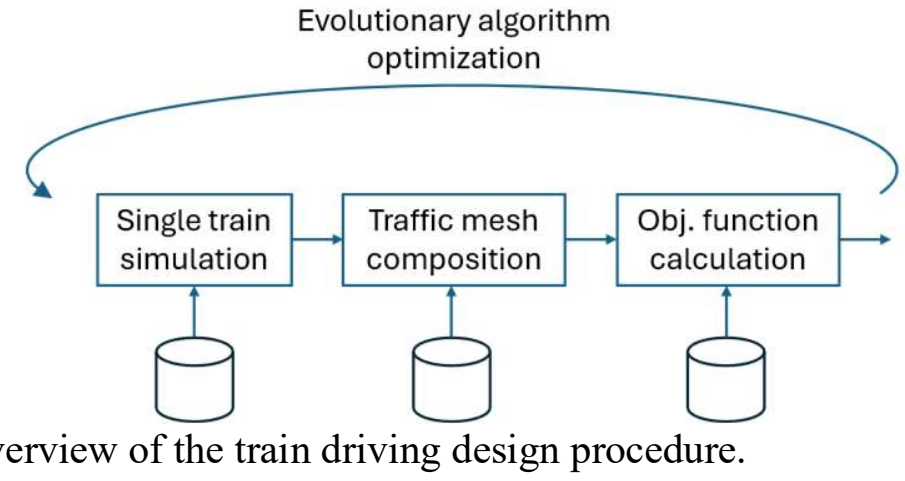

**Fig. 1.** Overview of the train driving design procedure.

The cost of the electricity includes both the cost of consumed energy and the cost of capacity utilization (i.e. the term to the maximum power allowed by the electricity supply cost). To the authors' best knowledge, this approach has not yet been developed in technical literature.

### *B. Single Train Simulation*

The single-train simulation module determines each train's movement, providing a given service defined by an origin, a destination, and a series of intermediate stops. This is done by integrating the differential equation given by Newton's second law (see Fig. 2), in the movement direction:

$$k_{inertial} \cdot m \cdot a = F_{trac} - F_w - F_{curv} - F_{res} \quad (1)$$

where $k_{inertial}$ is the coefficient of rotating inertial mass (typical values are between 1.05 and 1.15) of the train, $m$ is its mass, $a$ is its acceleration, $F_{trac}$ is the traction force, $F_w$ is the projection of the weight of the train in the direction of the movement, $F_{curv}$ is the resistant force due to the curves, $F_{res}$ is the running resistance force.

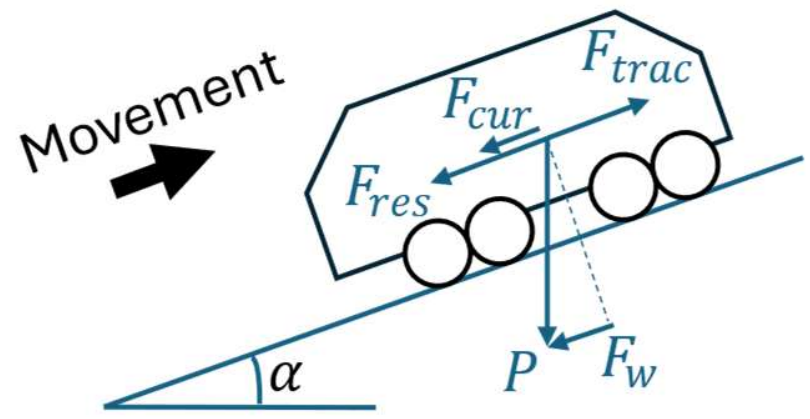

**Fig. 2.** Forces involved in the train movement

The force $F_w$ can be expressed as:

$$F_w = m \cdot g \cdot sin\alpha \quad (2)$$

where m is the mass of the train, and $g$ is the acceleration due to gravity (a constant value of 9.8 m/s$^2$ is assumed).

The force $F_{curv}$ is the resistance due to the track curves, and it is typically obtained by means of empirical expressions such as [16]:

$$F_{curv} = m \cdot \frac{k}{R_{curv}} \text{ (in N)} \quad (3)$$

where $k$ is an empirical value (in this work 600·g has been used) and $R_{curv}$ is the radius of curvature in the horizontal plane (in m).

The force $F_{res}$ is the running resisting force, and it is typically obtained by means of empirical expressions such as the Davis formula or similar [16][17]:

$$F_{res} = a + b \cdot v + c \cdot v^2 \quad (4)$$

where $v$ is the speed of the train, and a, b and c are coefficients that are obtained empirically for each type of train.

The driver of the train controls its movement by continuously $F_{trac}$ adjusting the traction equipment's force within its force limits and ensuring that the maximum speed allowed on the track and the maximum allowed acceleration are not exceeded.

Although many ways of driving a train for a given service are possible, the Minimum Train Driving (MTD) is usually taken as a reference. In MTD, the trains use these driving modes (see Fig. 3):

- Acceleration (referred to as ACC in the figure), subject to the train's maximum force limitation and to the comfort limitations,
- Cruising (referred to as CRU), in which the train moves at constant speed, and
- Braking (referred to as BR), normally at constant deceleration.

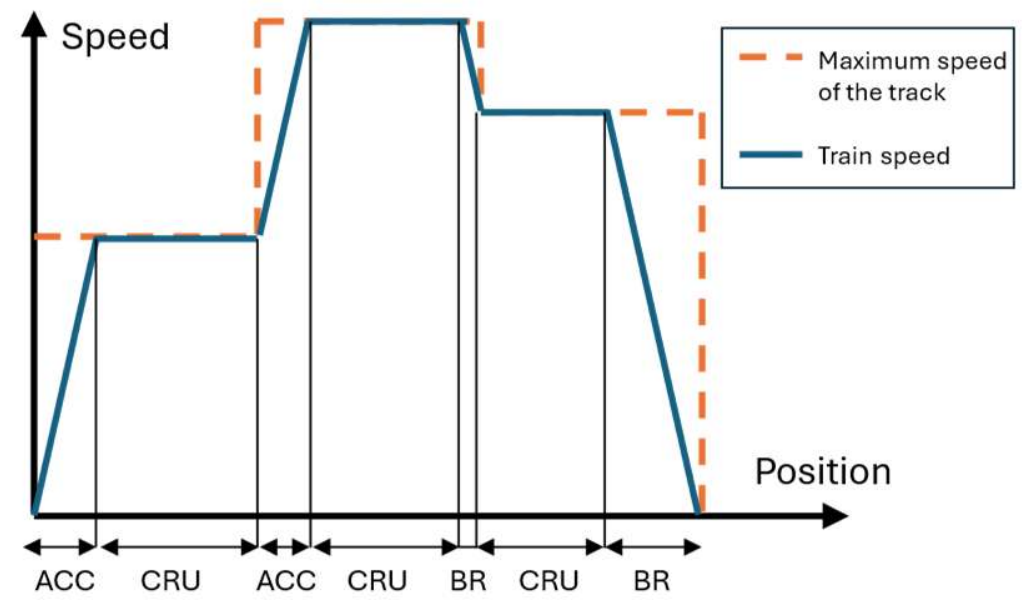

**Fig. 3.** MTD driving modes

Optimized driving usually consists of driving the train in such a way that the journey takes more time but, in exchange, a benefit is achieved (typically energy savings). In addition to the described driving modes, optimized driving can also include a fourth driving mode, coasting, in which the train moves only by inertia, without consuming any traction power. As shown in Fig. 4, coasting is used for different purposes such as an energy-efficient way of losing speed or for speed regulation.

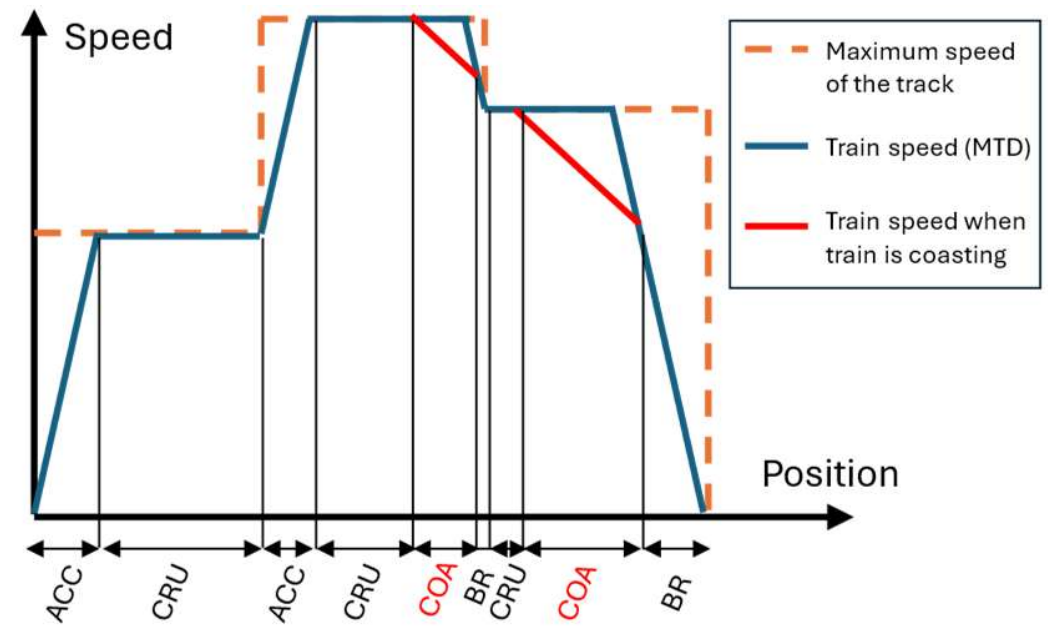

**Fig. 4.** Driving modes in optimized driving

Once the traction force $F_{traction}$ is known for every time step, the corresponding electric power consumption $P_{elec}$ can be calculated as:

$$P_{elec} = \begin{cases} \dfrac{F_{trac} \cdot v}{\eta_{elec \to mec}} + P_{anc} \ \ if \ F_{trac} > 0 \\ F_{trac} \cdot v \cdot \eta_{mec \to elec} + P_{anc} \ \ if \ F_{trac} < 0 \end{cases} \quad (5)$$

where $P_{elec}$ is the electrical power consumed by the train, $\eta_{elec \to mec}$ is the efficiency of the conversion of electrical into mechanical power when tractioning, $\eta_{mec \to elec}$ is the efficiency of the conversion of mechanical into electrical power when using regenerative braking, and $P_{anc}$ is the power consumed by the ancillary systems of the trains (lighting, air conditioning, etc.).

### *C. Traffic Mesh Construction*

The traffic model of this work is composed of the superposition of several periodic traffic meshes, one for each train service included in the analysis. The traffic mesh is constructed by copying and time-shifting the single-train simulation (see Fig. 5), according to the starting time $T_{ini,serv,s}$ and the period $T_{serv,s}$ of each service s.

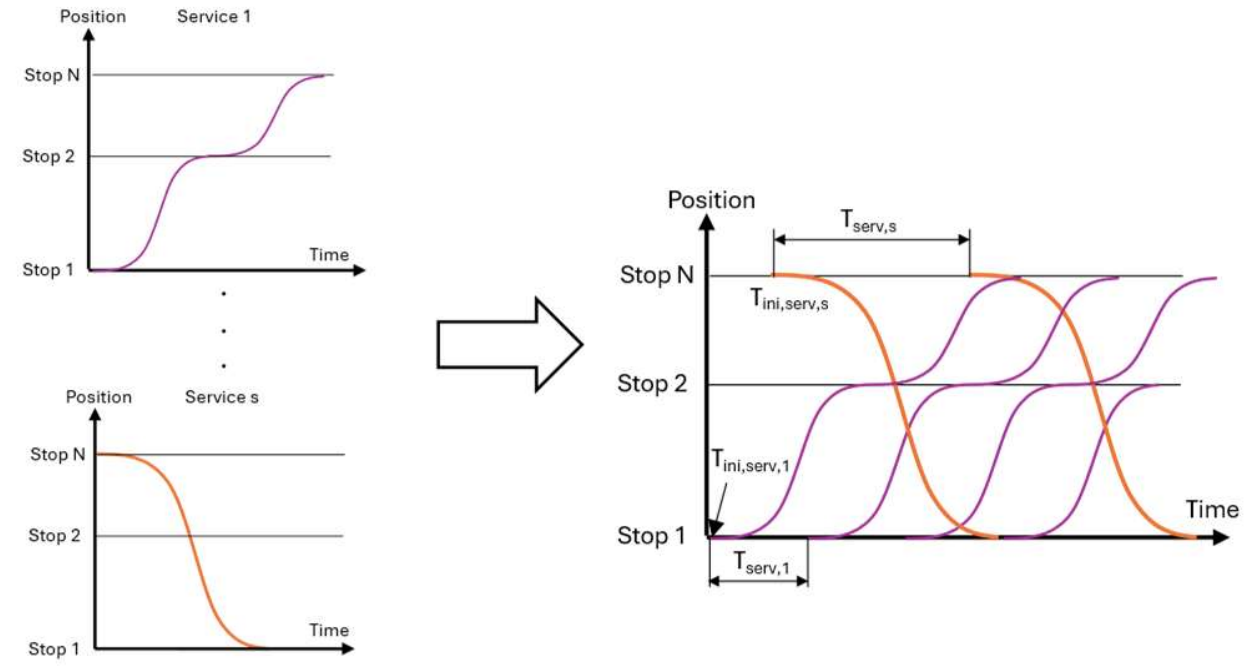

**Fig. 5.** Construction of the overall traffic mesh

The traffic model supports any mesh that can be formed as a set of periodic meshes, defined by the list of stops, the train type, the parameters $T_{ini,serv,s}$ and $T_{serv,s}$.

### *D. Optimization*

Taking as a reference the MTD of each service, the optimization intends to minimize the total cost of the electricity used by the trains, including both the cost of the electricity (cost per MWh) and the cost of the power allocated by the supply contract (cost per MW), by using the flexibility that exists in terms of journey duration [18].

The optimization has been implemented using the Evolutionary Computation Framework (ECF), a C++ framework intended for application of evolutionary computation [19] that supports metaheuristics algorithms such as steady state tournament, generational roulette-wheel, elimination, particle swarm optimization (PSO), differential evolution (DE), genetic annealing, artificial bee colony (ABC), clonal selection (CLONALG), immune optimization (optIA), evolution strategy, and random search. The evaluation of candidate solutions uses the traffic simulation described above, which is also implemented in C++.

The optimization searches for the way of driving each train service that minimizes the electricity cost, while not exceeding the maximum admissible time margin. To do so, the objective function is calculated as:

$$OF = COST_{Pmax} + COST_{Ener} + PEN_{delay} \quad (6)$$

with

$$COST_{Pmax} = \sum_{ez} C_{P,ez} \cdot \max\left(P_{SS,ez}(t), t\right), \quad (7)$$

$$COST_{Ener} = \sum_{ez} \sum_{t} C_{E,ez} \cdot P_{SS,ez}(t) \cdot \Delta t, \text{ and} \quad (8)$$

$$PEN_{delay} = \sum_{s} C_{delay} \cdot EX_{del}(s). \quad (9)$$

where $C_{P,ez}$ is the cost of the maximum power established in the supply contract for the electrical zone *ez*, $P_{SS,ez}(t)$ is the total power supplied at the traction substation of the electrical zone *ez* at instant *t*, $C_{E,ez}$ is the energy cost in the electrical zone *ez*, $\Delta t$ is the time step, $C_{delay}$ is the penalty established for exceeding the maximum allowed flexibility, and $EX_{del}(s)$ is the time exceeding the maxim allowed duration of service *s*.

To explore ways of driving that reduce energy and power consumption of the trains, the following actions are combined: (i) reducing the maximum speed, (ii) reducing the maximum traction force, (iii) reducing the maximum acceleration, and (iv) adding coasting sections. To ensure these actions can be applied where they have a bigger impact, the railway is divided into optimization sections, which are created as follows.

#### *D.1 Maximum speed optimization*

The optimization sections are created by dividing the line so that each section belongs to at most one maximum-speed section of the track (see Fig. 6).

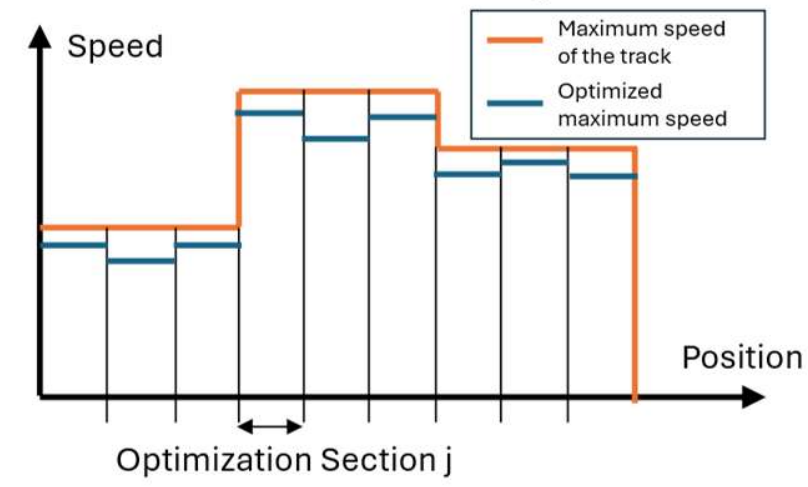

**Fig. 6.** Optimization sections for maximum speed and force

For each optimization section j, variables $Factor_{MaxSpeed}(j)$ are defined within the range [0,1], and new operational speed limits are introduced to the single-train simulation:

$$v \leq Factor_{MaxSpeed}(j) \cdot V_{max,track}(j) \quad (10)$$

where $V_{max,track}(j)$ is the maximum speed of the track at the location of the optimization section j.

#### *D.2 Maximum traction force optimization*

The optimization of the traction force consists of reducing the maximum traction force (i.e to make accelerations smoother) in sections where this leads to an improvement in terms of the objective function. To do so, the same optimization sections considered for maximum speed optimizations are used, and variables $Factor_{MaxForce}(j)$ are defined within the range [0,1] and operational traction force limits are introduced to the single-train simulation:

$$F_{traction} \leq Factor_{MaxForce}(j) \cdot F_{traction,max}(tt, v) \quad (11)$$

where $F_{traction,max}(tt, v)$ is the maximum value $F_{traction}$ of a train type tt at a speed v.

### D.3 Maximum acceleration optimization

Another way of making the movement smoother is by limiting the acceleration using the variables $Factor_{MaxAccel}(\mathrm{sl})$, which are defined within the range [0,1] for each maximum speed section sl defined for the track. Operational acceleration limits are also added to the single-train simulation:

$$a \leq Factor_{MaxForce}(j) \cdot a_{max} \quad (12)$$

where $a_{max}$ is the maximum admissible acceleration.

### D.4 Coasting section optimization

Finally, coasting sections are introduced for every speed reduction existing in the line (see Fig. 7). For each coasting section cs, its end corresponds to the speed reduction and its length $L_{coast}(cs)$ depends on the optimization variables $Factor_{CoastLength}(\mathrm{cs})$, defined within the range [0,1].

$$L_{coast}(cs) = Factor_{CoastLength}(\mathrm{cs}) \cdot L_{coasting,max} \quad (13)$$

where $L_{coasting,max}$ is the maximum coasting distance preset for the optimization.

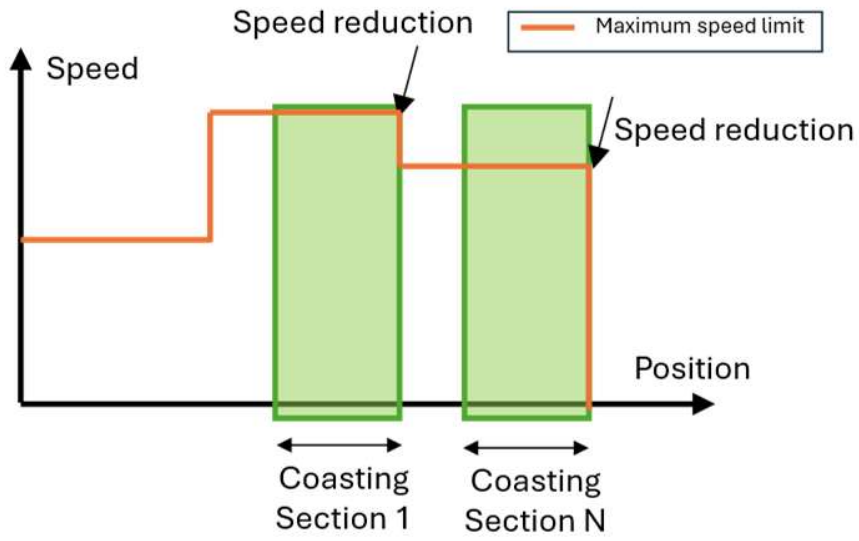

**Fig. 7.** Optimization coasting sections

## IV. Case Study

To evaluate the performance of the proposed optimization procedure, a 445 km long section of the high-speed line (HSL) between Madrid and Barcelona (Spain) is considered.

### A. Description of the Madrid-Lleida Section

The outline of the considered section is shown in Fig. 8, including the two electrical zones used to establish contracted power and energy prices.

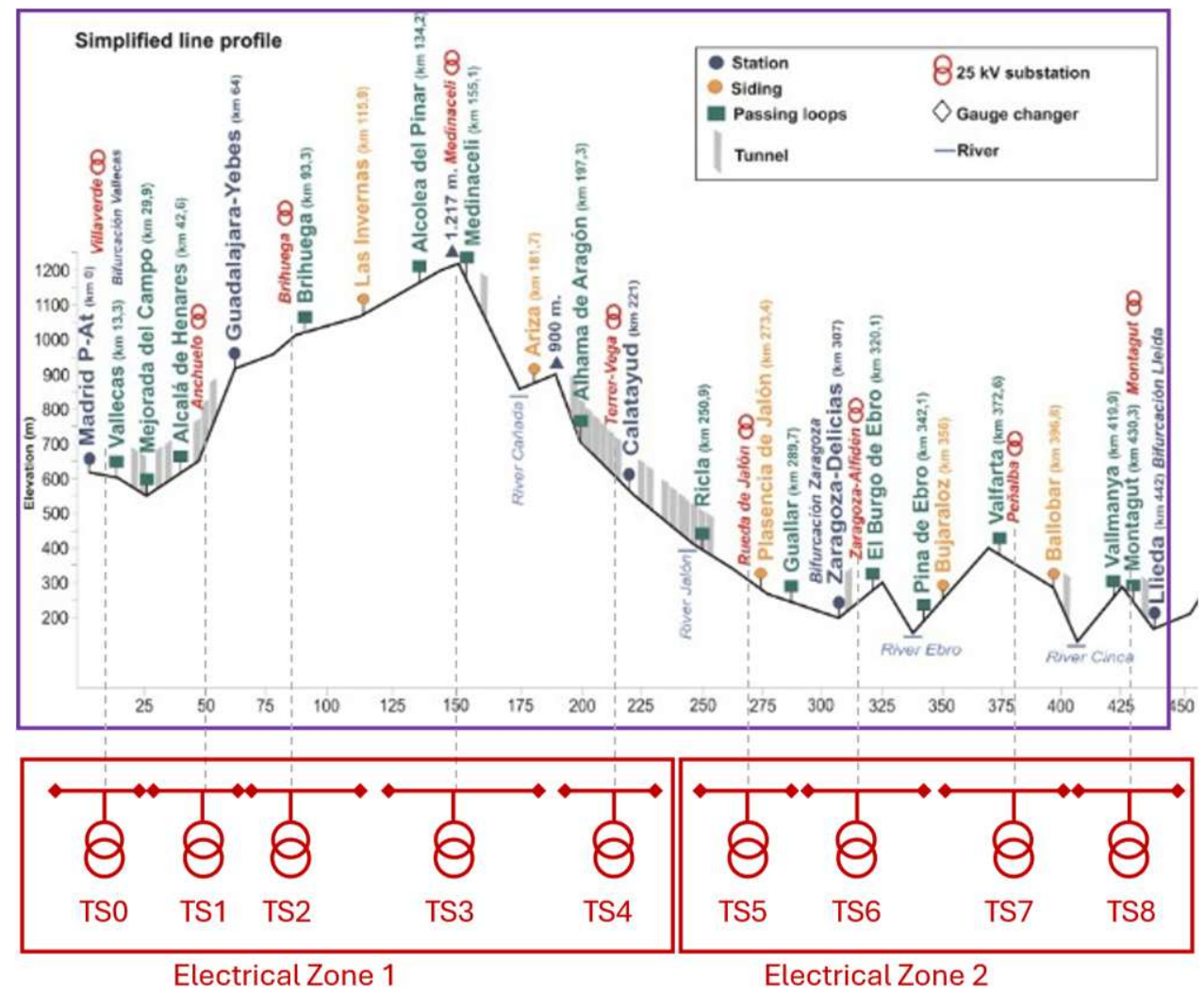

**Fig. 8.** Characteristics of the considered HSL section.

The traffic used in this case study consists of train circulations from Madrid to Lleida, with an intermediate stop in Zaragoza, in both directions (Madrid-Lleida and Lleida-Madrid), every 10 minutes. The duration of all the stops is 5 minutes, identically in MTD and optimized driving. Maximum acceleration and deceleration in MTD is 0,7 m/s$^2$.

All train circulations use series 103 trains, whose most relevant characteristics are shown in Fig. 9.

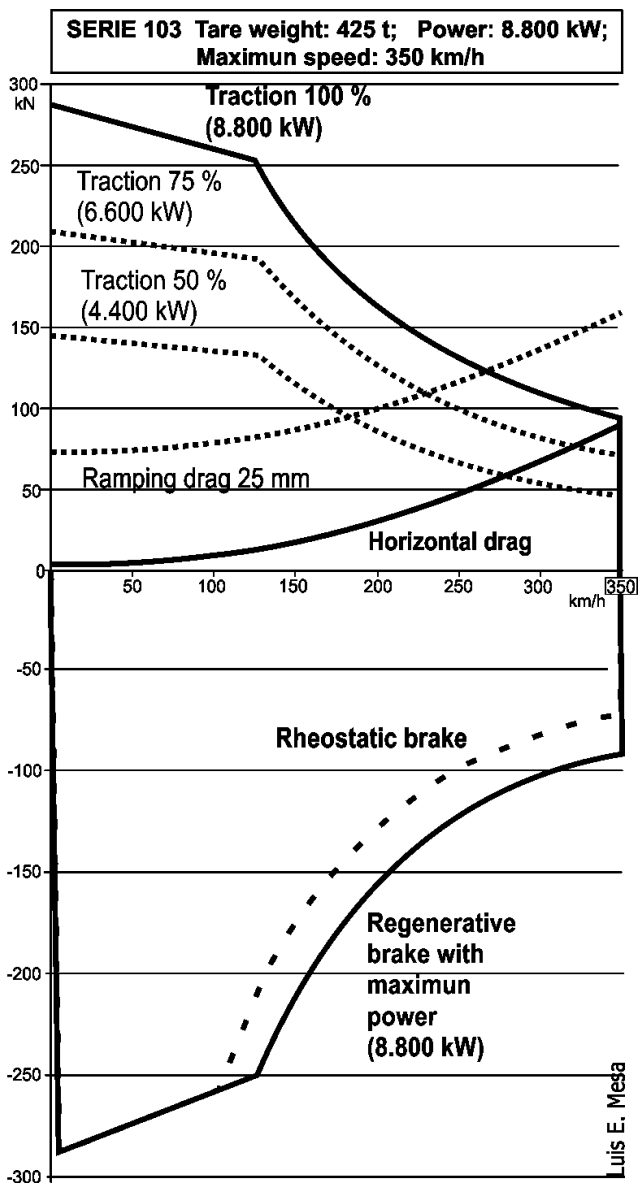

**Fig. 9.** Characteristics of the S-103 trains.

For all the cases analyzed, the time step has been set to $\Delta t$=4 s. Also, since traffic is periodic, the simulation time has been set to one period ($T_{sim}$ =10 minutes).

The trip duration using MTD is 5712 s from Madrid to Lleida and 5736 s from Lleida to Madrid (stop durations not included).

### *B. Case 1: Two Energy Price Sections, No Power Capacity Costs*

The electricity costs used in this case, which differ between the two electrical zones, are shown Table I. The time margin is 10 minutes.

TABLE I. OBJECTIVE FUNCTION PARAMETERS. CASE 1.

| | Electrical zone 1 | Electrical zone 2 |
|---|---|---|
| $C_{P,ez}$ (c€/kW) per Tsim | 0 | 0 |
| $C_{E,ez}$ (€/MWh) | 6 | 12 |
| $C_{delay}$ (€/s) | 1000 | 1000 |

The optimization tool is configured to use the steady-state tournament algorithm. Fig. 10 shows the values of the optimization variables obtained after a 1-hour execution.

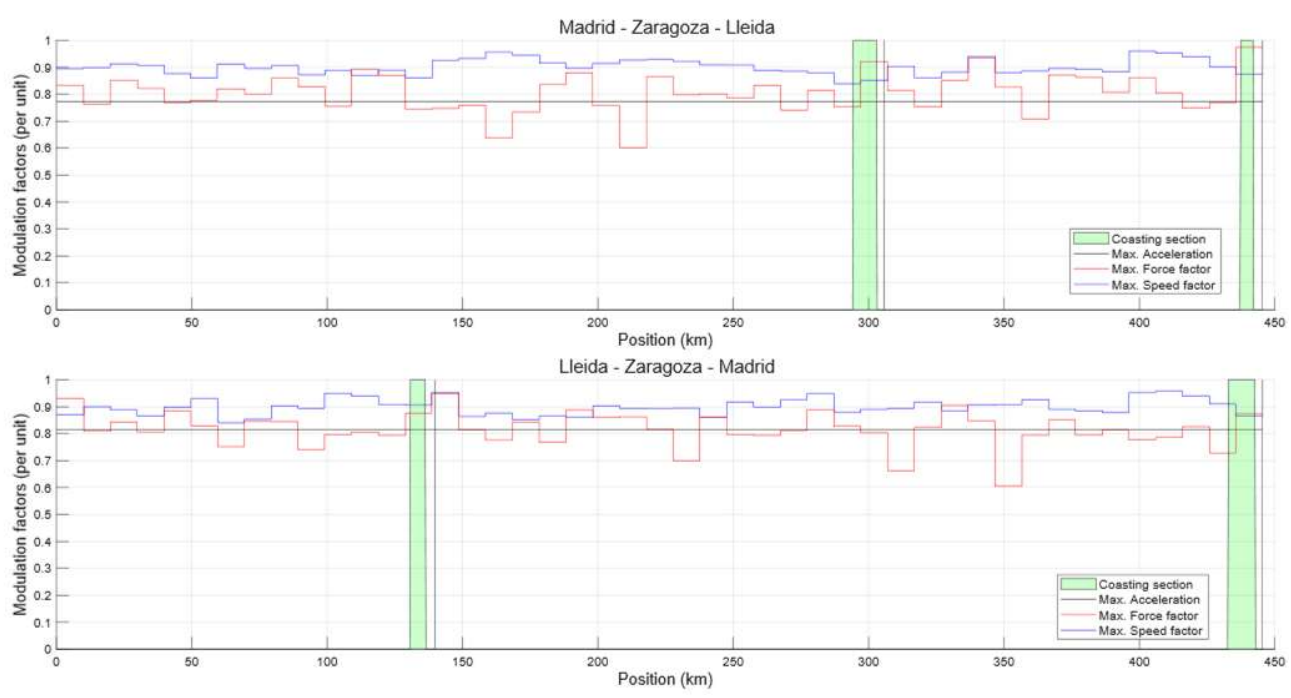

**Fig. 10.** Values of the optimization variables. Case 1.

For this optimal solution, the maximum power peak and the energy supplied by each substation are shown in Table II and Table III, respectively. The energy consumption reduction achieved by the optimization is 12% and 16% in electrical zones 1 and 2, respectively. Since the energy cost in electrical zone 2 is higher, the optimization tends to save more energy in this zone.

Also, although power peaks shaving is not targeted in Case 1 (its assigned cost is zero), the energy consumption indirectly reduces the maximum power peaks by 6%-25%, depending on the electrical zone.

TABLE II. MAXIMUM SUPPLIED POWER PEAKS. CASE 1.

| Electrical Zone | Substation | Maximum power peak, per substation (MW) | | | Sum of the maximum power peaks, per electrical zone (MW) | | |
|---|---|---|---|---|---|---|---|
| | | MTD | Optimized | Variation | MTD | Optimized | Variation |
| EZ1 | 1 | 21.06 | 17.41 | -17% | 114.87 | 108.23 | -6% |
| | 2 | 19.37 | 17.76 | -8% | | | |
| | 3 | 20.32 | 16.88 | -17% | | | |
| | 4 | 33.34 | 32.51 | -2% | | | |
| | 5 | 20.78 | 23.66 | 14% | | | |
| EZ2 | 6 | 36.49 | 25.89 | -29% | 122.48 | 91.62 | -25% |
| | 7 | 26.23 | 21.94 | -16% | | | |
| | 8 | 38.70 | 25.12 | -35% | | | |
| | 9 | 21.06 | 18.68 | -11% | | | |

TABLE III. TOTAL SUPPLIED ENERGY. CASE 1.

| Electrical Zone | Substation | Consumed energy per period Tsim, per substation (MWh) | | | Total consumed energy per period Tsim, | | |
|---|---|---|---|---|---|---|---|
| | | E_MTD_kWh | E_Opt_kWh | Variation | MTD | Optimized | Variation |
| EZ1 | 1 | 1111 | 878 | -21% | 10034 | 8815 | -12% |
| | 2 | 1774 | 1599 | -10% | | | |
| | 3 | 2033 | 1808 | -11% | | | |
| | 4 | 3016 | 2631 | -13% | | | |
| | 5 | 2100 | 1900 | -10% | | | |
| EZ2 | 6 | 2285 | 1956 | -14% | 9020 | 7570 | -16% |
| | 7 | 2436 | 1975 | -19% | | | |
| | 8 | 2620 | 2265 | -14% | | | |
| | 9 | 1678 | 1375 | -18% | | | |

Table IV shows the improvements achieved by the optimization tool compared to the MTD operation. By using the available time margin, the optimal solution can reduce the total energy cost by 15%.

TABLE IV. OBJECTIVE FUNCTION COMPONENTS. CASE 1.

| | MTD | Optimal driving | Variation |
|---|---|---|---|
| Trip duration (s): | | | |
| Madrid/Lleida | 5712 | 6312 | 600 |
| Lleida/Madrid | 5736 | 6336 | 600 |
| Terms of the objective function (€): | | | |
| $COST_{Pmax}$ | 0 | 0 | 0% |
| $COST_{Ener}$ | 1684 | 1437 | -15% |
| Objective Function | 1684 | 1437 | -15% |

Fig. 11 compares the speed of the train services in the original MTD and the optimized driving, in both directions.

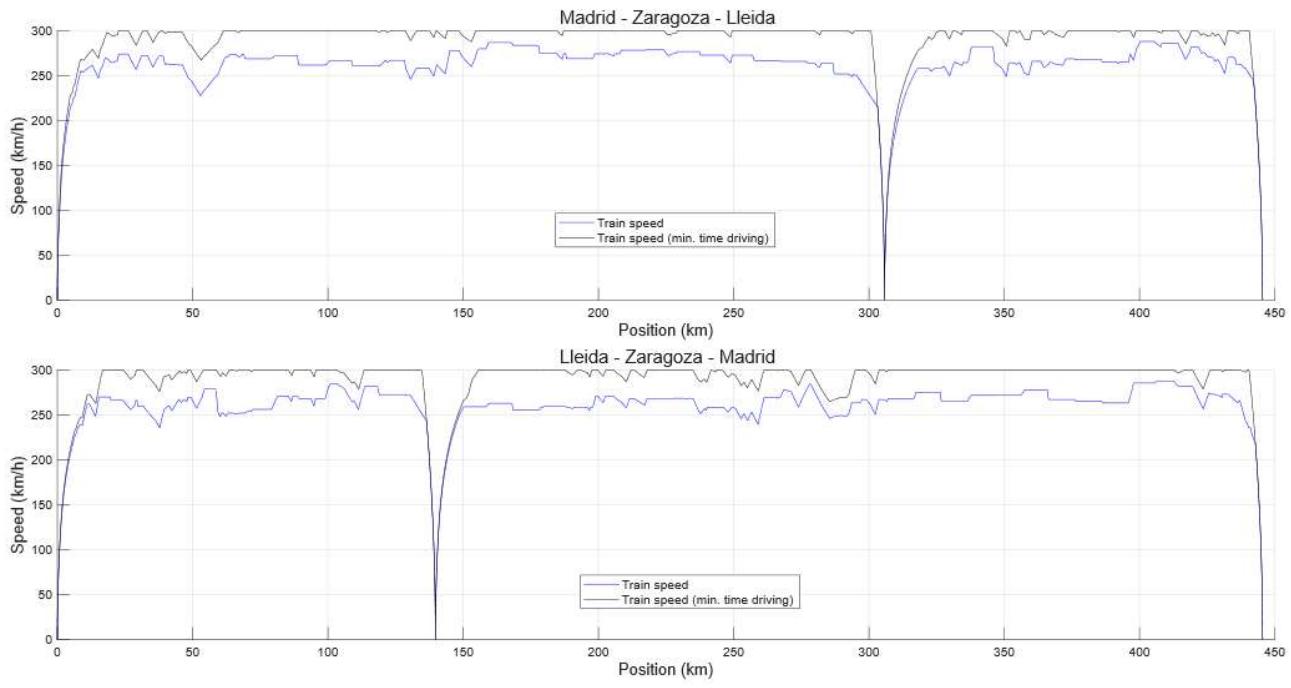

**Fig. 11.** Comparison of speed profile. Case 1.

Fig. 12 compares the traction forces optimized traffic with those of the original MTD operation. Please note that the area below these curves is proportional to the traction energy the train consumes, making it easy to visualize the energy savings achieved by the optimization.

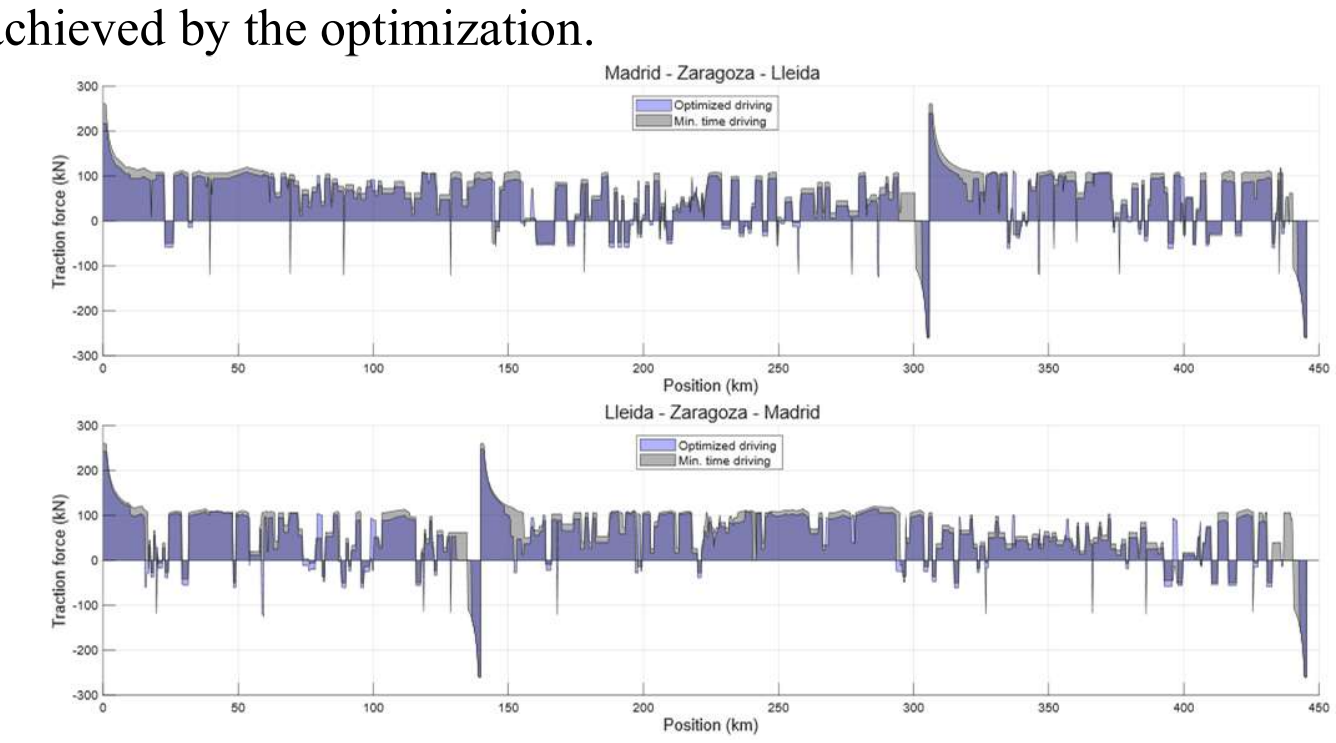

**Fig. 12.** Comparison of traction force. Case 1.

Fig. 13 compares the total power consumption at the traction substations of the two electrical zones considered. Please note that the area below these curves corresponds to the supplied energy, making it easy to visualize the energy savings achieved by the optimization. The power peaks can also be identified by the maximum values of these data series.

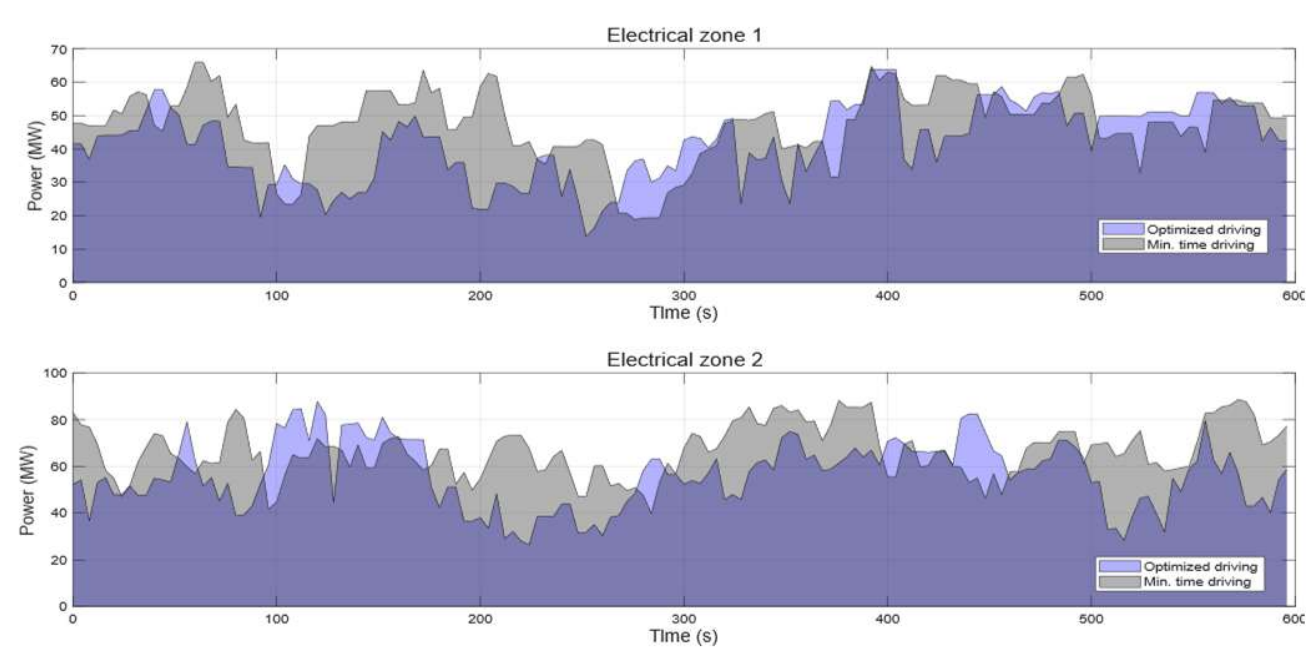

**Fig. 13.** Comparison of power consumption. Case 1.

Finally, a sensitivity analysis has been carried out to estimate the variation of the objective function when the time margins change (see Fig. 14).

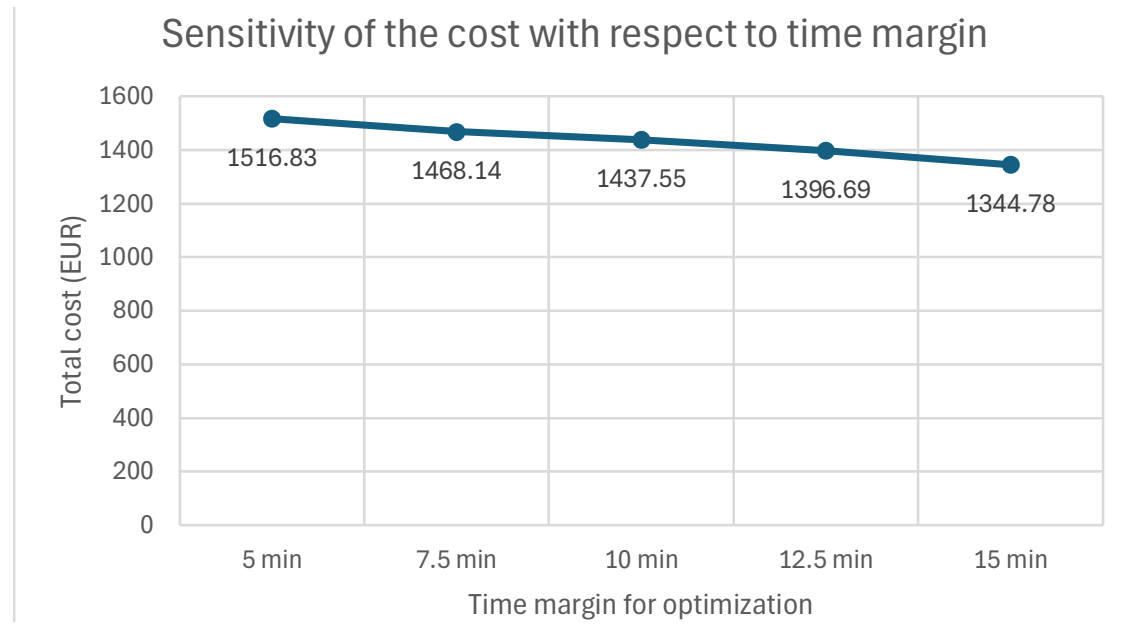

**Fig. 14.** Sensitivity of total cost vs time margin. Case 1.

*C. Case 2: Two Power-Capacity Cost Sections, No Energy Costs*

The electricity costs used in this case, which differ between the two electrical zones, are shown in Table V. The time margin is 10 minutes.

TABLE V. OBJECTIVE FUNCTION PARAMETERS. CASE 2.

| | Electrical zone 1 | Electrical zone 2 |
|---|---|---|
| $C_{P,ez}$ (c€/kW) per Tsim | 0.09 | 0.18 |
| $C_{E,ez}$ (€/MWh) | 0 | 0 |
| $C_{delay}$ (€/s) | 1000 | 1000 |

The optimization tool is configured to use the steady-state tournament algorithm. Fig. 15 shows the values of the optimization variables obtained after a 1-hour execution.

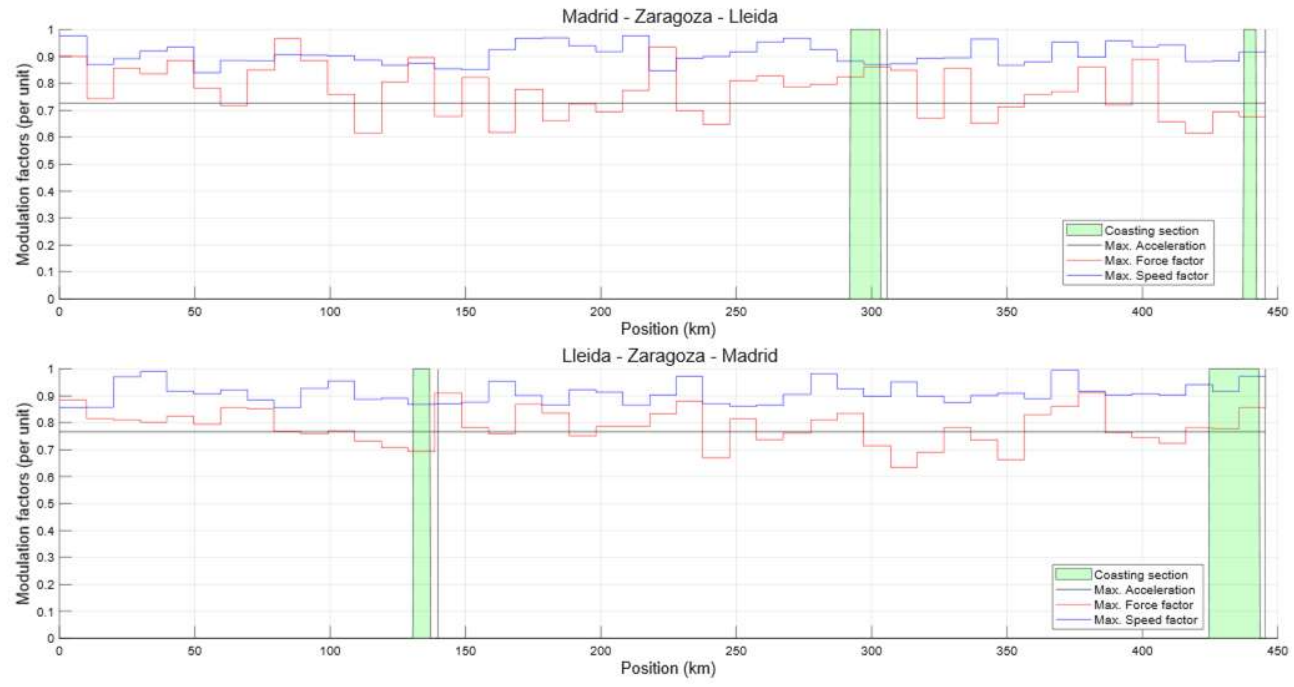

**Fig. 15.** Values of the optimization variables. Case 2.

For this optimal solution, the maximum power peak and the energy supplied by each substation are shown in Table V and Table VI, respectively. The power peak reduction achieved by the optimization is 11% and 33% in electrical zones 1 and 2, respectively. Since the maximum power cost in electrical zone 2 is higher, the optimization tends to find solutions that shave power peaks more aggressively in this zone.

It should be noted that, although energy minimization is not the objective in Case 1 (its assigned cost is zero), energy consumption is indirectly reduced by 13% and 15%.

TABLE VI. MAXIMUM SUPPLIED POWER PEAKS. CASE 2.

| Electrical Zone | Substation | Maximum power peak, per substation (MW) | | | Sum of the maximum power peaks, per electrical zone (MW) | | |
|---|---|---|---|---|---|---|---|
| | | MTD | Optimized | Variation | MTD | Optimized | Variation |
| EZ1 | 1 | 21.06 | 16.55 | -21% | 114.87 | 102.66 | -11% |
| | 2 | 19.37 | 16.01 | -17% | | | |
| | 3 | 20.32 | 17.31 | -15% | | | |
| | 4 | 33.34 | 27.88 | -16% | | | |
| | 5 | 20.78 | 24.92 | 20% | | | |
| EZ2 | 6 | 36.49 | 19.22 | -47% | 122.48 | 81.70 | -33% |
| | 7 | 26.23 | 21.04 | -20% | | | |
| | 8 | 38.70 | 24.31 | -37% | | | |
| | 9 | 21.06 | 17.14 | -19% | | | |

TABLE VII. TOTAL SUPPLIED ENERGY. CASE 2.

| Electrical Zone | Substation | Consumed energy per period Tsim, per substation (MWh) | | | Total consumed energy per period Tsim, | | |
|---|---|---|---|---|---|---|---|
| | | E_MTD_kWh | E_Opt_kWh | Variation | MTD | Optimized | Variation |
| EZ1 | 1 | 1111 | 896 | -19% | 10034 | 8752 | -13% |
| | 2 | 1774 | 1562 | -12% | | | |
| | 3 | 2033 | 1842 | -9% | | | |
| | 4 | 3016 | 2638 | -13% | | | |
| | 5 | 2100 | 1814 | -14% | | | |
| EZ2 | 6 | 2285 | 1867 | -18% | 9020 | 7626 | -15% |
| | 7 | 2436 | 2081 | -15% | | | |
| | 8 | 2620 | 2276 | -13% | | | |
| | 9 | 1678 | 1402 | -16% | | | |

Table VIII shows the improvements achieved by the optimization tool compared to the MTD operation. By using the available time margin, the optimal solution can reduce the total maximum power cost by 26%.

TABLE VIII. OBJECTIVE FUNCTION COMPONENTS. CASE 2.

| | MTD | Optimal driving | Variation |
|---|---|---|---|
| Trip duration (s): | | | |
| Madrid/Lleida | 5712 | 6300 | 588 |
| Lleida/Madrid | 5736 | 6332 | 596 |
| Terms of the objective function (€): | | | |
| $COST_{Pmax}$ | 323.8 | 239.5 | -26% |
| $COST_{Ener}$ | 0 | 0 | 0% |
| Objective Function | 324 | 239 | -26% |

Fig. 16, Fig. 17, and Fig. 18 compare the speed of the train services, the traction force, and the total power consumption of the optimized traffic to those of the original MTD operation.

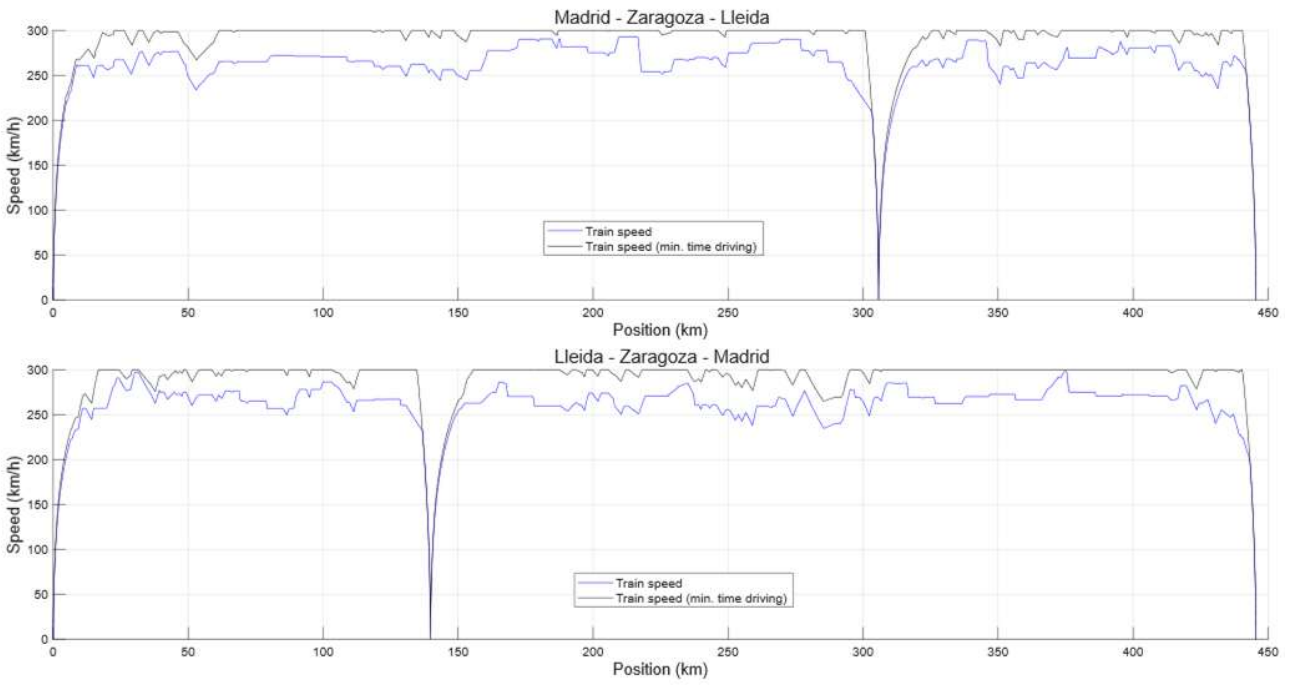

**Fig. 16.** Comparison of speed profile. Case 2.

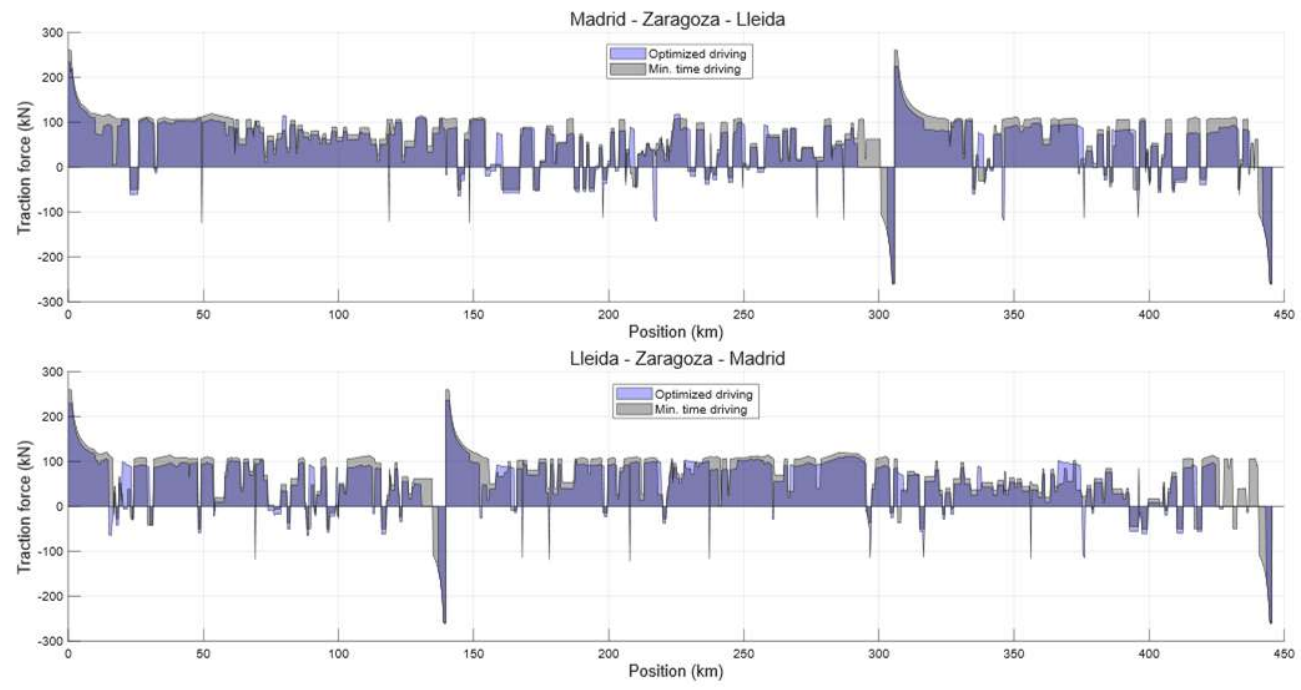

**Fig. 17.** Comparison of traction force. Case 2.

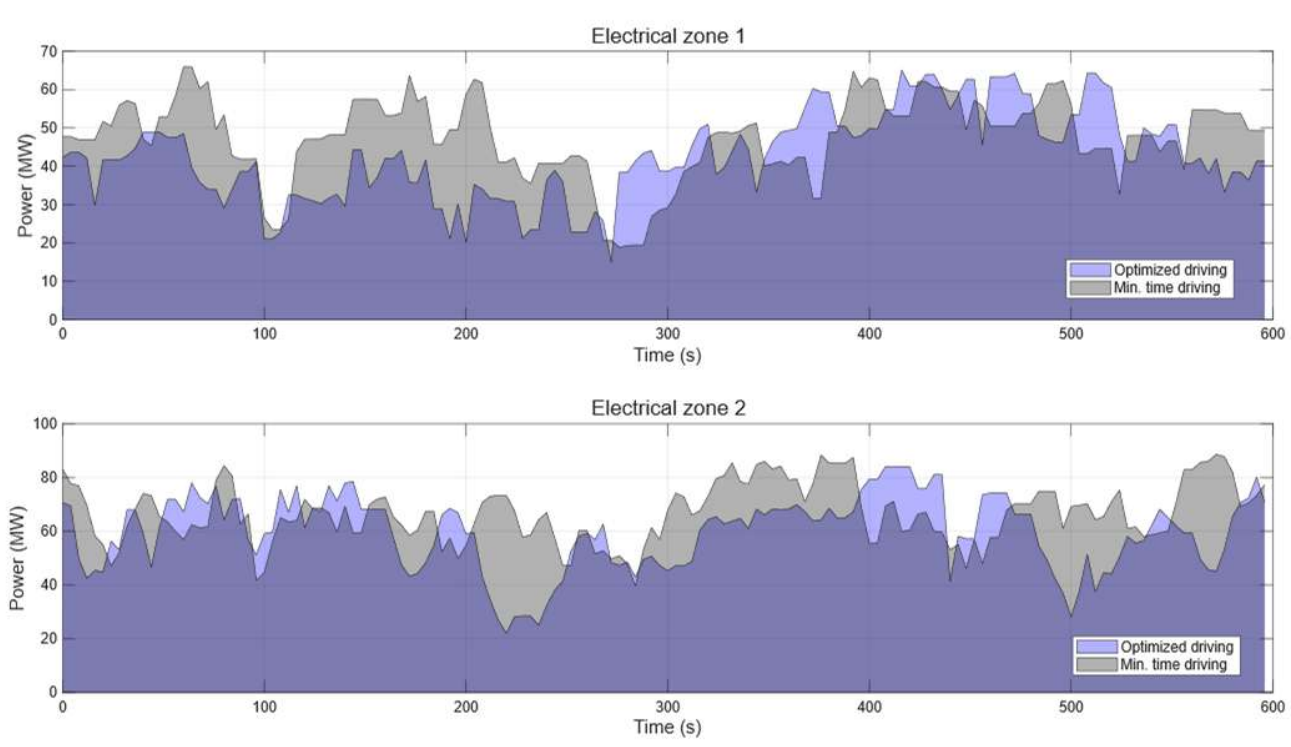

**Fig. 18.** Comparison of power consumption. Case 2.

Finally, a sensitivity analysis has been conducted to estimate the variation in the objective function as the time margins change (see Fig. 19).

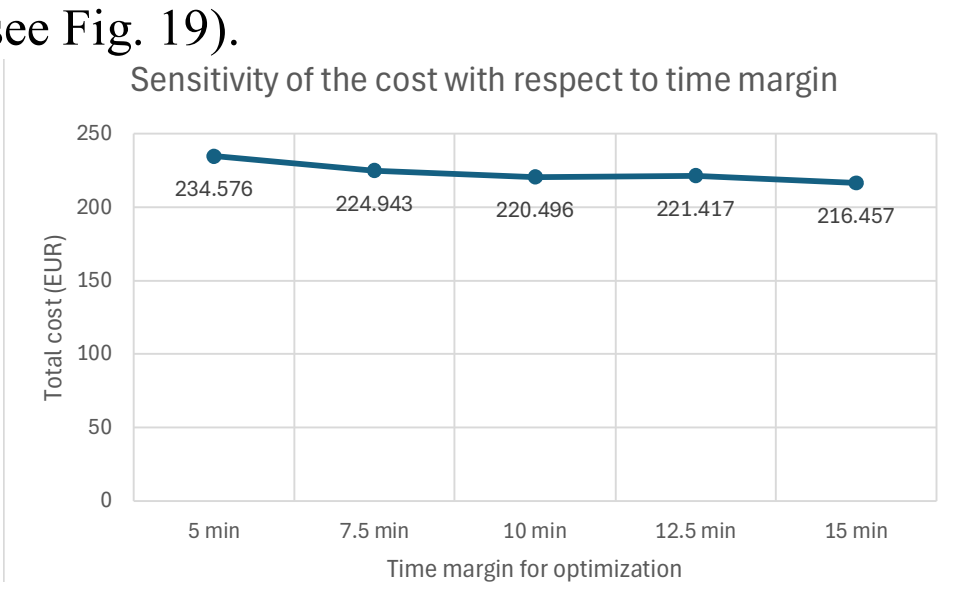

**Fig. 19.** Sensitivity of total cost vs time margin. Case 2.

## *F. Case 3: Two Energy and Power Cost Sections*

The electricity costs used in this case, which differ between the two electrical zones, are shown in TABLE IX. The time margin is 10 minutes.

TABLE IX. OBJECTIVE FUNCTION PARAMETERS. CASE 3.

| | Electrical zone 1 | Electrical zone 2 |
|---|---|---|
| $C_{P,ez}$ (c€/kW) per Tsim | 0.09 | 0.18 |
| $C_{E,ez}$ (€/MWh) | 6 | 12 |
| $C_{delay}$ (€/s) | 1000 | 1000 |

The optimization tool is configured to use the steady-state tournament algorithm. Fig. 20 shows the values of the optimization variables obtained after a 1-hour execution.

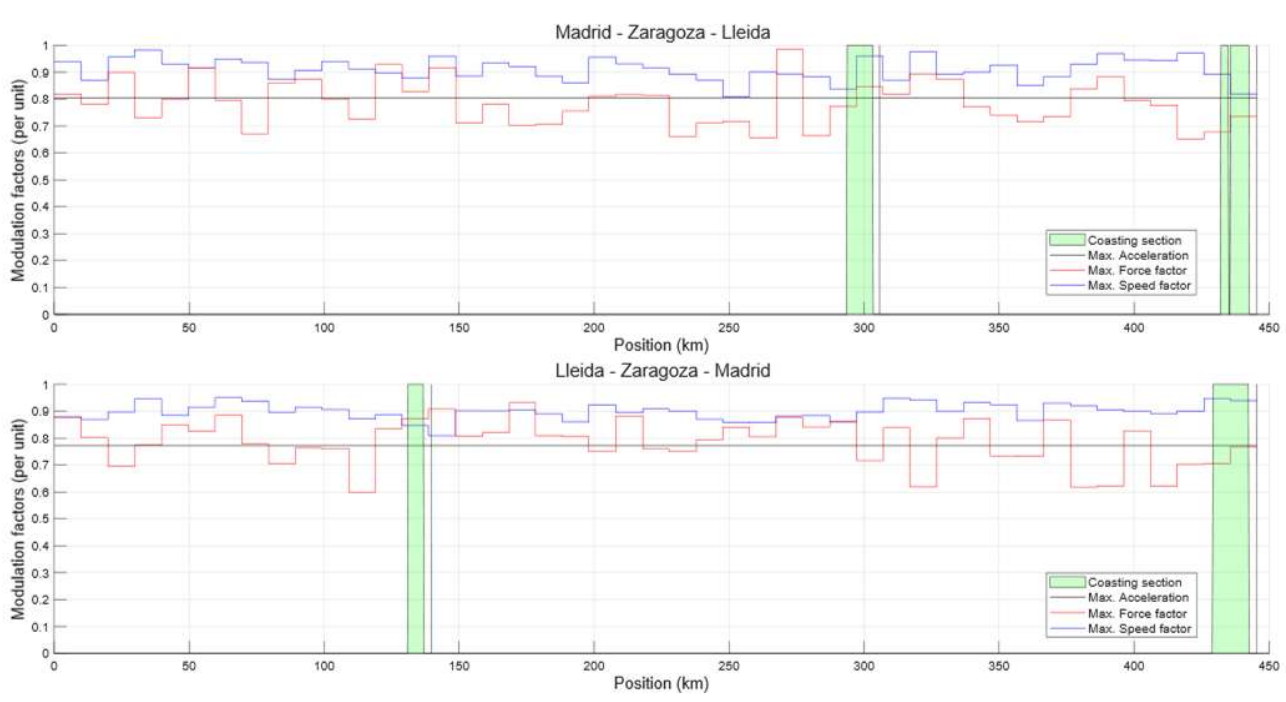

**Fig. 20.** Values of the optimization variables. Case 3.

For this optimal solution, the maximum power peak and the energy supplied by each substation are shown in Table X and

Table XI, respectively. The energy consumption reduction achieved by the optimization is 10% and 11% in electrical zones 1 and 2, respectively. The power peak reduction achieved by the optimization is 23% and 36% in electrical zones 1 and 2, respectively. Since the electricity costs are higher in electrical zone 2, the optimization tends to find solutions that shave power peaks and reduce energy consumption more aggressively in this zone.

TABLE X. MAXIMUM SUPPLIED POWER PEAKS. CASE 2.

| Electrical Zone | Substation | Maximum power peak, per substation (MW) | | | Sum of the maximum power peaks, per electrical zone (MW) | | |
|---|---|---|---|---|---|---|---|
| | | MTD | Optimized | Variation | MTD | Optimized | Variation |
| EZ1 | 1 | 21.06 | 10.13 | -52% | 114.87 | 88.41 | -23% |
| | 2 | 19.37 | 16.00 | -17% | | | |
| | 3 | 20.32 | 16.25 | -20% | | | |
| | 4 | 33.34 | 25.98 | -22% | | | |
| | 5 | 20.78 | 20.05 | -3% | | | |
| EZ2 | 6 | 36.49 | 19.51 | -47% | 122.48 | 78.29 | -36% |
| | 7 | 26.23 | 19.31 | -26% | | | |
| | 8 | 38.70 | 23.36 | -40% | | | |
| | 9 | 21.06 | 16.12 | -23% | | | |

TABLE XI. TOTAL SUPPLIED ENERGY. CASE 3.

| Electrical Zone | Substation | Consumed energy per period Tsim, per substation (MWh) | | | Total consumed energy per period Tsim, | | |
|---|---|---|---|---|---|---|---|
| | | E_MTD_kWh | E_Opt_kWh | Variation | MTD | Optimized | Variation |
| EZ1 | 1 | 1111 | 828 | -26% | 10034 | 9013 | -10% |
| | 2 | 1774 | 1599 | -10% | | | |
| | 3 | 2033 | 1893 | -7% | | | |
| | 4 | 3016 | 2797 | -7% | | | |
| | 5 | 2100 | 1896 | -10% | | | |
| EZ2 | 6 | 2285 | 1976 | -14% | 9020 | 8025 | -11% |
| | 7 | 2436 | 2125 | -13% | | | |
| | 8 | 2620 | 2398 | -8% | | | |
| | 9 | 1678 | 1526 | -9% | | | |

Table XII shows the improvements achieved by the optimization tool compared to the MTD operation. By using the available time margin, the optimal solution can reduce the total electricity cost by 14%. Compared to case 1, a lower reduction in energy is observed, but in exchange, a higher reduction in the power peaks. Compared to case 2, a higher reduction in the power peaks is observed, which shows how much the minimization of the power peaks and the energy consumption are interrelated.

TABLE XII. OBJECTIVE FUNCTION COMPONENTS. CASE 3.

| | MTD | Optimal driving | Variation |
|---|---|---|---|
| Trip duration (s): | | | |
| Madrid/Lleida | 5712 | 6312 | 600 |
| Lleida/Madrid | 5736 | 6336 | 600 |
| Terms of the objective function (€): | | | |
| $COST_{Pmax}$ | 323.8 | 220.5 | -32% |
| $COST_{Ener}$ | 1684 | 1504 | -11% |
| Objective Function | 2008 | 1724 | -14% |

Fig. 21, Fig. 22, and Fig. 23 compare the speed of the train services, the traction force, and the total power consumption of the optimized traffic with those of the original MTD operation.

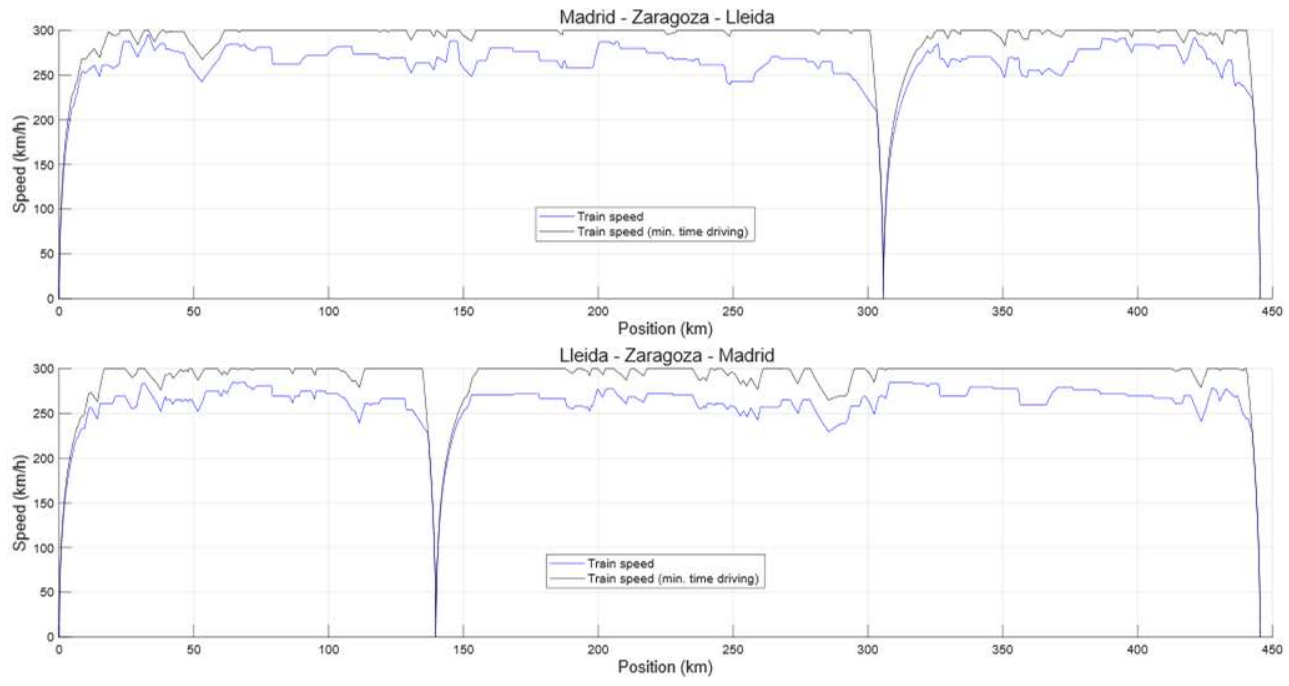

**Fig. 21.** Comparison of speed profile. Case 3.

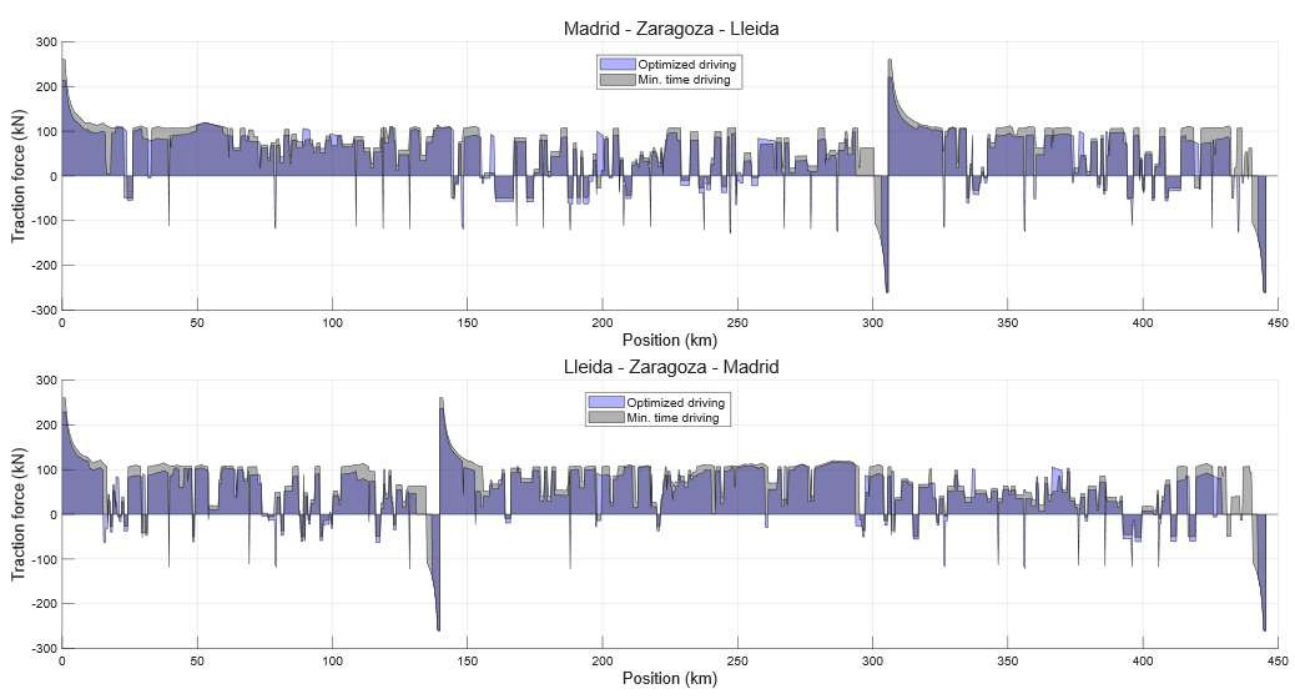

**Fig. 22.** Comparison of traction force. Case 3.

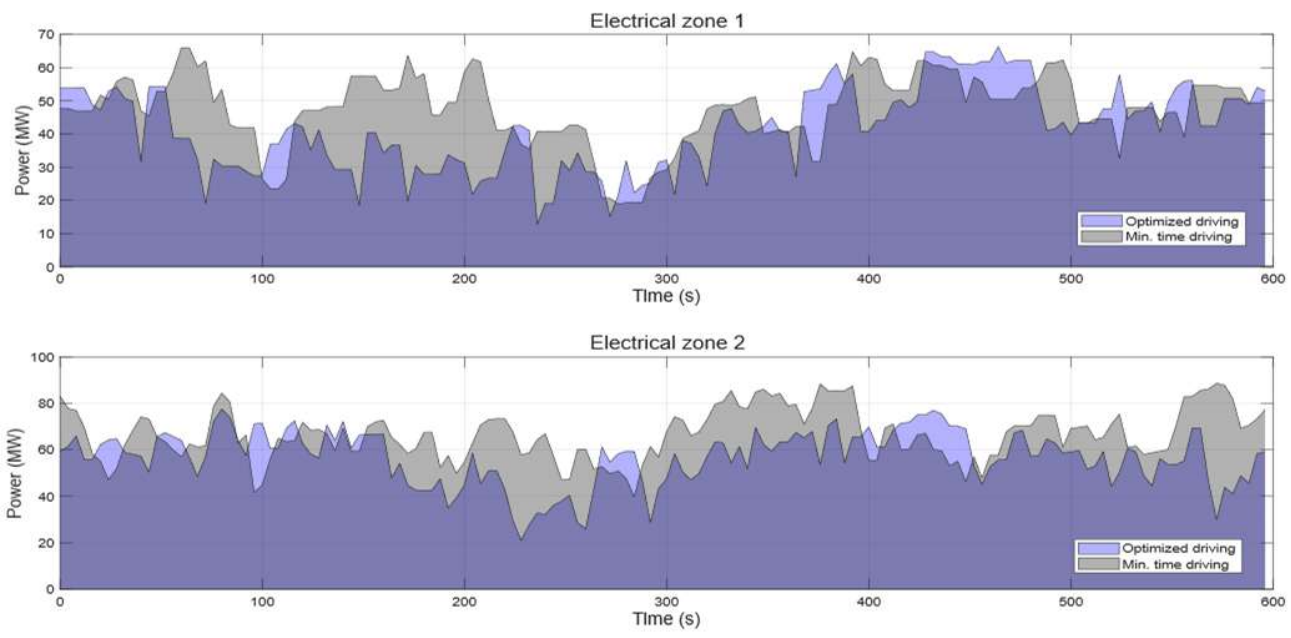

**Fig. 23.** Comparison of power consumption. Case 3.

Finally, a sensitivity analysis has been carried out to estimate the variation of the objective function when the time margins change (see Fig. 24).

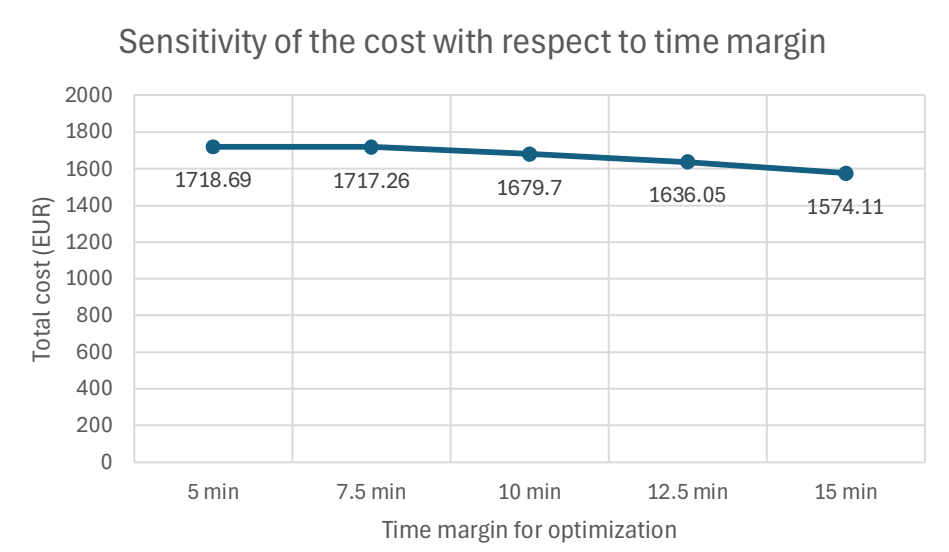

**Fig. 24.** Sensitivity of total cost vs time margin. Case 3.

## V. Conclusions

This paper has presented a procedure to optimize train operations to minimize the cost of supplying electrical energy, including not only energy consumption costs but also the cost of power capacity utilization, using a zone-dependent pricing scheme. This optimization process can be applied to complex periodic traffic meshes, composed of multiple services and train types.

This optimization has been implemented in C++ and tested on a 445 km-long section of the high-speed line between Madrid and Barcelona (Spain), using Siemens S-103 trains and different electrical power and energy scenarios. In the study cases, reductions of around 14% in the total electricity bill and 32% in the power-dependent term have been achieved.